\documentclass[letterpaper, 10 pt, conference]{ieeeconf}  

\IEEEoverridecommandlockouts                              

\title{\LARGE \bf
DBPF: A Framework for Efficient and Robust Dynamic Bin-Picking}
\author{Yichuan Li$^{1}$,~\IEEEmembership{Student member,~IEEE}, Junkai Zhao$^{2}$, Yixiao Li$^{2}$, \\Zheng Wu$^{3}$, Rui Cao$^{1}$, Masayoshi Tomizuka$^{3}$,~\IEEEmembership{Fellow,~IEEE}, Yun-Hui Liu$^{1,2}$,~\IEEEmembership{Fellow,~IEEE}
\thanks{*This work is supported by the InnoHK of the Government of the Hong Kong Special Administrative Region via the Hong Kong Centre for Logistics Robotics. \textcolor{black}{The work is also supported in part by RGC via grant 14207119.} Corresponding to Yun-Hui Liu (e-mail: yhliu@mae.cuhk.edu.hk)}%
\thanks{$^{1}$Yichuan Li, Rui Cao, Yun-Hui Liu are with the T Stone Robotics Institute, the Department of Mechanical and Automation Engineering, the Chinese University of Hong Kong (e-mail: yichuan.li@link.cuhk.edu.hk; rcao@mae.cuhk.edu.hk; )}
\thanks{$^{2}$Junkai Zhao, Yixiao Li and Yun-Hui Liu are with the Hong Kong Centre for Logistics Robotics (HKCLR)
(e-mail: zhaojunkai1515@gmail.com; \protect\\yixili@student.ethz.ch)}
\thanks{$^{3}$Masayoshi Tomizuka and Zheng Wu are with the Department of Mechanical Engineering, 
		University of California at Berkeley (e-mail: tomizuka@berkeley.edu; zheng\_wu@berkeley.edu)}
}
\usepackage{graphicx}
\usepackage{xcolor}
\usepackage{mathtools}
\usepackage{algorithm}
\usepackage{algorithmicx}
\usepackage{algpseudocode}
\usepackage{amsmath}
\usepackage[inkscapelatex=false]{svg}
\usepackage{lettrine}
\usepackage{multirow}
\usepackage{booktabs}
\usepackage{cite}
\usepackage{setspace}
\begin{document}

\maketitle
\thispagestyle{empty}
\pagestyle{empty}

\begin{abstract}
Efficiency and reliability are critical in robotic bin-picking as they directly impact the productivity of automated industrial processes. However, traditional approaches, demanding static objects and fixed collisions, lead to deployment limitations, operational inefficiencies, and process unreliability.
This paper introduces a Dynamic Bin-Picking Framework (DBPF) that challenges traditional static assumptions. The DBPF endows the robot with the reactivity to pick multiple moving arbitrary objects while avoiding dynamic obstacles, such as the moving bin.
Combined with scene-level pose generation, the proposed pose selection metric leverages the Tendency-Aware Manipulability Network optimizing suction pose determination. 
Heuristic task-specific designs like velocity-matching, dynamic obstacle avoidance, and the resight policy, enhance the picking success rate and reliability.
Empirical experiments demonstrate the importance of these components. Our method achieves an average 84\% success rate, surpassing the 60\% of the most comparable baseline, crucially, with zero collisions. Further evaluations under diverse dynamic scenarios showcase DBPF's robust performance in dynamic bin-picking. Results suggest that our framework offers a promising solution for efficient and reliable robotic bin-picking under dynamics.

\end{abstract}
\vspace{0.5em}
\section{Introduction}
\lettrine[lines=2]{B}{in-picking}, a fundamental problem in robotics, involves a robot manipulator retrieving items from a bin. 
The existing bin-picking methods\cite{levine2018learning,dong2019ppr,yang2021robi} widely rely on open-loop workflow, i.e., Sense-Plan-Act (SPA). This approach depends on a limited number of perceptions prior to robot execution and has demonstrated practicality in static scenarios where both objects and the environment remain stationary.
However, many real-world applications in manufacturing, logistics, and retail industries often entail dynamic scenarios. A typical situation might require a robot manipulator tasked with picking target objects from a bin in transit, moved by a belt conveyor or a mobile robot. Under such circumstances, the static assumptions of traditional bin-picking operations become inflexible and overly restrictive, leading to operational inefficiencies and unreliability.
\begin{figure}
  \vspace{0.25em}
  \centering
  \includegraphics[width=0.47\textwidth]{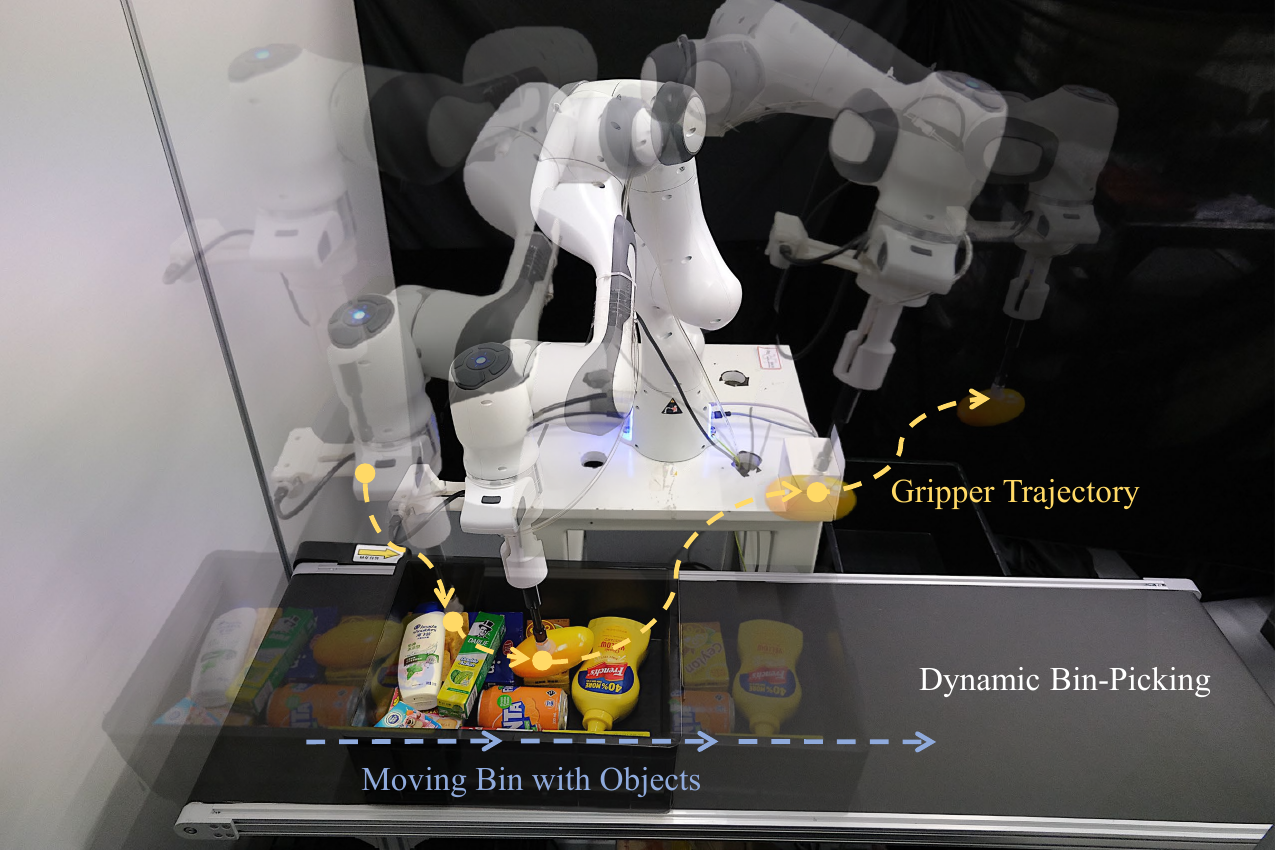}
  \vspace{-0.5ex}
  \caption{Dynamic bin-picking: The robot picks up a single object among a cluster of arbitrary objects stacked in a moving bin on a belt conveyor. Our novel framework enables the robot to achieve dynamic bin-picking, ensuring collision-free and adaptive motion. Our method outperforms traditional bin-picking methods in terms of time efficiency, showcasing exceptional effectiveness and reliability in dynamic scenarios.}
  \label{fig1}
  \vspace{-1.75em}
\end{figure}

Current bin-picking research primarily focuses on static environments, based on perception techniques such as 3D object localization \cite{cao2023two}, 6D pose estimation \cite{yang2021probabilistic}\cite{li2022sim}, and grasping planning \cite{zhang2022learning}\cite{domae2014fast} to process sensor data and generate effective picking poses for stationary items. However, these approaches are susceptible to unexpected external disturbances on initially stationary scenes once the robot commences a picking motion \cite{bohg2013data}\cite{roy2016pose}. Furthermore, if these methods were applied to pick objects from a moving bin, the effort for frequent halting and restarting of the bin could significantly degrade picking efficiency and lead to unnecessary energy consumption. In contrast, picking objects directly from a moving bin, i.e., dynamic bin-picking (Fig. \ref{fig1}), could reduce time overheads, enhance productivity, and save energy. Therefore, it is worth exploring dynamic bin-picking, which challenges the static assumptions of traditional bin-picking scenarios. The real-time generation of a feasible picking pose is essential to realize dynamic bin-picking. Some previous studies have addressed moving object picking\cite{de2021dual,akinola2021dynamic,islam2021provably} by integrating 6D Pose estimation and grasping planning. However, these strategies are primarily aimed at a single-moving object of limited categories and fall short when dealing with multiple-moving arbitrary objects inherent in dynamic bin-picking. As the quantity and diversity of objects increase, the preparatory work for the 6D pose estimation and instance segmentation models becomes more burdensome. Besides, the moving bin itself poses a significant dynamic obstacle, and the consideration of bin collisions, which are often overlooked\cite{ichnowski2020gomp}\cite{fang2023anygrasp}, is crucial. Dynamic bin-picking involving moving objects and dynamic obstacles necessitates a motion planning framework capable of real-time adaptation to changing objectives and constraints.

This work advances dynamic bin-picking by relaxing static assumptions common in traditional approaches. Our contributions are:
(1) A novel framework utilizing horizon-based discrete trajectory optimization for rarely explored dynamic bin-picking, enabling efficient collision-free robot motion to pick moving objects from moving or disturbed bins.
(2) The introduction of the pose selection metric, including the Tendency-Aware Manipulability Network (TAMN), with a combination of the scene-level suction pose generation, facilitating optimal pose determination for multiple moving arbitrary objects.
(3) Heuristic, task-specific designs incorporated into optimization objectives, constraints, and task planning model, enhancing picking success rate and reliability in practice.
(4) Empirical evidence from real-world experiments showcasing the effectiveness and reliability of our framework for dynamic bin-picking, outperforming competitive baseline methods.

\vspace{-0.75em}
\section{RELATED WORK}
\textbf{Static Bin-Picking}:
The conventional static bin-picking method typically adheres to a Sense-Plan-Act framework. The Sense module analyzes sensor data to determine the picking poses of the objects, the Plan module devises an offline motion trajectory, and the Act module executes this predetermined trajectory to retrieve the target. The primary task in static bin-picking lies in determining a reliable picking pose for object~\cite{bui2023deep}\cite{mahler2019learning}. Numerous studies have delved into deep learning-based object picking pose estimation~\cite{li2022sim,buchholz2013efficient,dong2019ppr,kleeberger2019large}. Zeng et al.\cite{zeng2022robotic} proposed grasping primitives applicable to diverse objects, utilizing fully convolutional networks to infer grasping affordances. Similarly, we propose leveraging learning-based methods for picking pose generation, but we further extend beyond by incorporating Tendency-Aware Manipulability into consideration for moving objects. Ichnowski et al.~\cite{ichnowski2020gomp} proposed a rapid bin-picking method using sequential quadratic programming (SQP) to optimize the robot's motion. However, this approach might not be applicable in dynamic environments where bin movements introduce additional complexity and unpredictability that are not considered in a static SQP framework. 

\textbf{Moving object picking}: In the process of moving object picking, the robot needs to continuously track the object and reason about picking pose, meanwhile adjusting the arm motion accordingly. Akinola el al.~\cite{akinola2021dynamic} proposed a dynamic object grasping framework where the appropriate grasp is filtered from reachability and motion-awareness and the arm motion is generated by PRM in a seeding way. 
However, their framework only specializes in grasping a single object moving on a belt conveyor and excludes environmental collision. In contrast, our work deals with more complex scenarios where the robot picks an object from a clutter of arbitrary objects in a moving bin.
Burgess-Limerick et al.~\cite{burgess2022dgbench} use the image-based visual servoing control to grasp objects with unpredictable movements. However, the top-down grasp pose lacks generality across various object types. 
\textcolor{black}{Further, ~\cite{de2021dual} proposes a dynamic grasp re-ranking metric to select the best 6D grasp pose for the moving object. 
Position-based visual servoing is adopted but often overlooks collision risks when reaching objects. Unlike most previous studies using two-fingered grippers, we employ suction cups to better suit industrial settings. We present a novel framework to pick multiple moving objects while ensuring the avoidance of moving obstacles, i.e., the moving bin.}

\vspace{-0.25em}
\section{DYNAMIC BIN-PICKING FRAMEWORK}
\begin{figure*}[htbp]
\vspace{0.5em} 
\centering
   \makebox[\textwidth][c]{\includegraphics[width=0.94\textwidth]{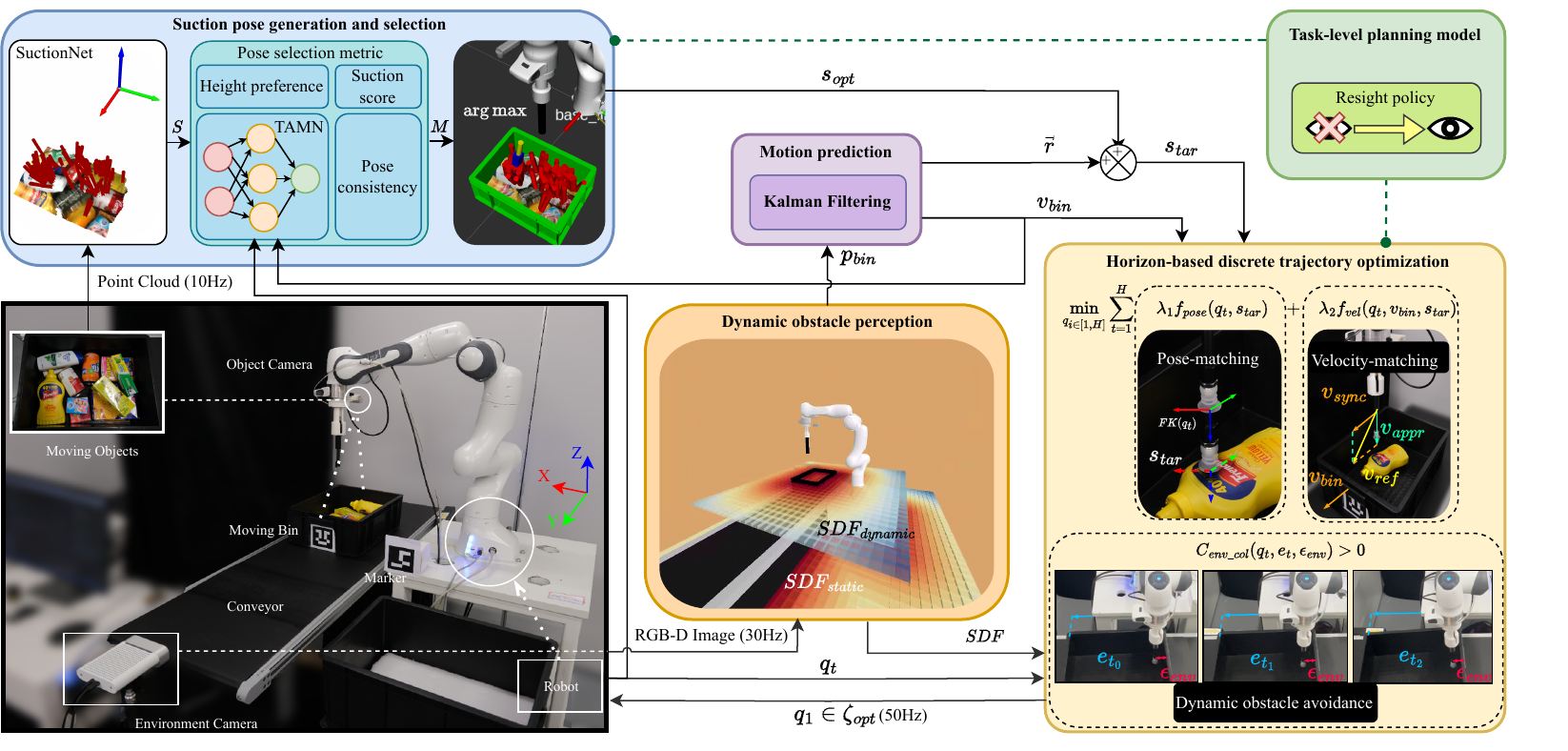}}
   \vspace{-1.5em} 
  \caption{Framework overview: The DBPF is implemented in a fully closed-loop manner to facilitate the robot's reactivity in dynamic bin-picking. An eye-in-hand object camera captures the point cloud of moving objects within the bin at 10Hz. With the suction pose generation and selection, a set of candidate suction poses $S$ are generated, and an optimal pose $s_{opt}$ (yellow cylinder) is selected considering factors like the Tendency-Aware Manipulability (through TAMN), pose consistency, height preference, and suction score. Motion prediction forecasts bin's displacement $\vec r$ to advance the $s_{opt}$ as $s_{tar}$ and offers bin velocity $v_{bin}$. An eye-to-hand environment camera obtains RGB-D images at 30Hz, and the dynamic obstacle perception module detects the pose of the moving bin $p_{bin}$. Two Signed Distance Functions $SDF_{static}$ and $SDF_{dynamic}$ are maintained actively corresponding to environmental collisions. Horizon-based discrete trajectory optimization solves the optimal trajectory $\zeta_{opt}$ with objectives like pose-matching $f_{goal}$ and velocity-matching $f_{vel}$. Dynamic obstacle avoidance is realized via collision constraints. The first point $q_1$ of each optimal path $\zeta_{opt}$ is executed at 50Hz, and the robot's current joint position $q_t$ is constantly updated. Lastly, a task-level planning model integrates these modules tightly to effectively achieve dynamic bin-picking while preventing ``Poor Observations" through the resight policy.}
  \label{fig:framework}
  \vspace{-1.75em} 
\end{figure*}

\vspace{-0.25em}
In this section, we illustrate the proposed framework (As shown in Fig. \ref{fig:framework}) in detail, which is implemented in a fully closed-loop manner with several modules and a task-level planning model that allows the robot to actively adapt to changes during the dynamic bin-picking tasks.  

\vspace{-0.25em}
\subsection{Suction pose generation and selection}\label{sec:suc}
\textbf{Suction pose generation}: Pose candidates $S$ with strong suction potential are continuously generated from the post-processed point cloud, which keeps objects only. Given the inevitable occlusions for a third-view camera due to the presence of bin walls or the robot arm, we utilize the in-hand configuration.
Preliminary poses are randomly sampled as the normal on objects' surfaces and ranked using the SuctionNet (SN) considering factors of cup seal formation and wrench resistance \cite{cao2021suctionnet}.

\textbf{Suction pose selection}: 
Considering manipulability at the goal pose can help avoid potential singularity and increase the execution success rate. The ability to continue moving in a specific direction from a pose greatly influences the robot's continuous transition from the current pose along the predicted direction of motion. Therefore, we define the Tendency-Aware Manipulability (TAM) score for a given pose and its moving tendency. We employ the inverse condition number $C$ as a more robust and representative metric for manipulability than $|J^{\text{T}}J|$ (as per Togai \cite{togai1986application}), given end-effector pose $p$:
\begin{equation}\label{eq:inv_cond}
C(p) = 
\begin{cases} 
\frac{1}{Cond(J(q))}, & \text{if }\exists \text{ }q \in\text{ $IK(p)$} \\
-1, & \text{otherwise}
\end{cases}
\end{equation}
Where the $q$ is joint positions from Inverse Kinematics (IK) \textcolor{black}{and the $J$ is the kinematic Jacobian of $q$}. Furthermore, we propose the use of a quality index $I(q,n_p)$ (as in \cite{vahrenkamp2012manipulability}), which is the magnitude of the Jacobian's gradient with respect to a moving direction vector, $n_p$:
\begin{equation}\label{eq:quality_index}
I(q,n_p) = |J(q)^Tn_p|
\end{equation}
By combining the manipulability metric $C$ with this quality index, we derive the TAM score:
\begin{equation}\label{eq:tendency_manip}
T(p,n_p) = I(IK(p),n_p)C(p)
\end{equation}

However, performing online IK and TAM computations for a set of candidate suction poses can be time-consuming, which could compromise reactivity \cite{akinola2021dynamic}. To resolve this, we trained a Tendency-Aware Manipulability Network (TAMN) to infer TAM scores given 6D poses and their moving tendencies. TAMN was trained using a Multi-Layer Perceptron (MLP) network and optimized by minimizing the Mean Squared Error (MSE) loss, \textcolor{black}{with a learning rate of 0.0001 for 200 epochs. The training and validation loss are lower than 0.004.} \textcolor{black}{For training dataset, we randomly sampled 100,000 picking pose $p$ within the workspace, and calculated 128 TAM scores via Eq. 1-3 for each $p$ by partitioning the space into 128 possible directions $n_p$ uniformly.} The re-ranking score of a suction pose $s$ from a set of candidate suction poses $S$ can be computed as the pose selection metric $M$:
\vspace{-0.25em} 
\begin{equation}\label{eq:app_score}
\begin{aligned}
M(s) = w_{1}T(p_t,\frac{r(s)-r(p_t)}{|r(s)-r(p_t)|}) +w_{2}T(s,\frac{v_{ref}(s)}{|v_{ref}(s)|})\\
+w_{3}\frac{1}{pd(s,s_{t-1})}+w_{4}\frac{Z(s)}{\bar Z(S)}+w_{5}U(s)
\end{aligned}
\end{equation}
Where the $r(s)$ is the translation vector of $s$, the $p_t$ is the current end-effector pose derived from Forward Kinematics $FK(q_t)$ of the current joint position $q_t$, the $v_{ref}(s)$ from Eq.\ref{eq:vref} is the potential moving velocity upon reaching $s$, the $pd(s,s_{t-1})$ refer to Eq. \ref{eq:pose_dist} is the pose difference between $s$ and the previously determined pose $s_{t-1}$, the $z(s)$ is the height of $s$, the $\bar Z(S)$ is the average height of all candidates $S$, the $U(s)$ is the suction score from SN. The $M$ balances terms with weights $w_i$ for (1) goal pose accessibility, (2) goal pose TAM, (3) pose consistency, (4) height preference, and (5) suction score. Thus the optimal suction pose $s_{opt}$ with the highest $M(s)$ is determined from a pool of candidate suction poses $S$ as $s_{opt} = \arg\max M$.

\vspace{-0.25em} 
\subsection{Dynamic Obstacle Perception}\label{section:collision_detection}
Static collisions (e.g., table, conveyor) are configured before the task, while real-time obstacle perception is needed for dynamic collisions like the moving bin. The pose of the moving bin $p_{bin}$, represented by five cuboids, is detected using fiducial markers. The robot's links are shown as spheres. Both static and dynamic collisions are modeled as cuboids and spheres, allowing analytical computation of Signed Distance Functions ($SDF$) in the defined workspace. For computational efficiency, two $SDF$ are maintained: $SDF_{static}$ is computed once at the task's start, and $SDF_{dynamic}$ is updated with each new observation of dynamic collisions. These $SDF$ are used to calculate minimum distance vectors between robot link spheres and collisions in real-time.

\vspace{-0.25em} 
\subsection{Motion prediction}\label{sec:predict}
Due to the inevitable time cost for perception, the inferred suction pose will always be antiquated. Consequently, prediction of the future state of the moving objects is necessary. As we have the online 6D pose of the moving bin $p_{bin}$ refer to Section \ref{section:collision_detection}, and assume the objects and bin are relatively stationary, we employ Kalman Filtering (KF)~\cite{kalman1960new} to predict the bin's displacement vector $\vec{r}$ and velocity $v_{bin}$ using a linear motion model. Thus, the $s_{opt}$ is advanced to yield the target suction pose $s_{tar}$: 
\vspace{-0.5em} 
\begin{equation}\label{eq:tar_pred}
s_{tar}=s_{opt}+\vec{r}
\end{equation}

\subsection{Horizon-based discrete trajectory optimization} \label{section:RMP}
The primary objective of dynamic bin-picking is to facilitate robotic motion toward objects in a moving bin with constraints.
To encapsulate all kinds of requirements comprehensively and solve appropriate trajectories online, we formulate the problem through a horizon-based discrete trajectory optimization structure:
\vspace{-0.5em} 
\begin{equation}\label{eq:cost}
\min_{q_{i \in [1, H]}} \sum_{t=1}^{H}\lambda_1 f_{pose}(q_t, s_{tar}) + \lambda_2 f_{vel}(q_t, v_{bin}, s_{tar})
\end{equation}
\vspace{-1.5em} 
\begin{align}
\begin{split}\label{eq:bound}
\text{subject~to:} \quad l_i \leq q_i \leq u_i, \;\\\dot{l}_i \leq \dot{q}_i \leq \dot{u}_i, \;\ddot{l}_i \leq \ddot{q}_i \leq \ddot{u}_i, \; \forall i 
\end{split}
\end{align}
\vspace{-1em} 
\begin{equation}\label{eq:vel_inter}
q_{t+1}-q_t = \dot{q}_{t+1}dt
\end{equation}
\vspace{-1.5em} 
\begin{equation}\label{eq:acc_inter}
\dot{q}_{t+1}-\dot{q}_t =  \ddot{q}_{t+1}dt
\end{equation}
\vspace{-1.5em} 
\begin{equation}\label{eq:self_col}
C_{self\_col}(q_t,\epsilon_{self})>0
\end{equation}
\vspace{-1.5em} 
\begin{equation}\label{eq:env_col}
C_{env\_col}(q_t,e_t,\epsilon_{env})>0
\end{equation}
Where the Eq. \ref{eq:cost} are the objective functions, the $\lambda_1$ and  $\lambda_2$ are objective weights; Eq. \ref{eq:bound} set the joint limitations; Eq. \ref{eq:vel_inter} and Eq. \ref{eq:acc_inter} establish the correlation among joint position, velocity, and acceleration; Eq. \ref{eq:self_col} is the self-collision constraint; Eq. \ref{eq:env_col} is the env-collision constraint\textcolor{black}{; $H$, the horizons, reflect the future steps considered. Optimal $H$ balances computational efficiency with planning quality, ascertained empirically.}

\textbf{Pose-matching objective}: The pose difference between two Cartesian poses $p_a$ and $p_b$ are considered as in \cite{bhardwaj2022storm}:
\begin{equation}\label{eq:pose_dist}
\begin{aligned}
pd(p_a,p_b) = ||\alpha_1(I - \prescript{w}{}{R}_b^T \prescript{w}{}R_a )||_2 \\
+ ||\alpha_2(\prescript{w}{}R_b^T \prescript{w}{}d_a - \prescript{w}{}R_b^T \prescript{w}{}d_b)||_2
\end{aligned}
\end{equation}
Where the $R$ is the rotation and the $d$ is the translation of the pose. The $\alpha_1$ and $\alpha_2$ are weights to bias importance. Consequently, the Pose-matching objective $f_{pose}$ can be derived as:
\vspace{-0.75em}
\begin{equation}\label{eq:suc_goal}
f_{pose}(q_t, s_{tar}) = pd(FK(q_t),s_{tar})
\vspace{-0.25em}
\end{equation}
This formulation enables the robot to approach the target pose by minimizing the pose difference through optimization.
  
\textbf{Velocity-matching objective}: To improve the effectiveness of picking during the final approach, we aim to maintain a relative still between the end-effector and the moving objects in the direction of bin motion. We refer to this as the Velocity-matching objective, inspired by the V-bar maneuver detailed in \cite{markley2014fundamentals}, a fundamental strategy employed in space station docking.
Given the z-axis of a picking pose $s$ as $N_z(s)$, a speed variable $\alpha_v$ determines the desired approaching velocity component of the end-effector through $v_{appr}(s)=\alpha_v N_z(s)$.
The desired synchronizing velocity component of the end-effector follows $v_{sync}=v_{bin}$. Therefore, the reference velocity of end-effector $v_{ref}$ and velocity matching objective $f_{vel}$ are derived as:
\vspace{-0.5em} 
\begin{equation}\label{eq:vref}
v_{ref}(s) = v_{sync}+v_{appr}(s) = v_{bin} + \alpha_v N_z(s)
\end{equation}
\vspace{-1.25em} 
\begin{equation}\label{eq:velocity_cost}
f_{vel}(q_t, v_{bin},s_{tar}) = ||\dot{p}_t-v_{ref}(s_{tar})||_1
\end{equation}
Where the $\dot{p}_t$ is the current end-effector velocity calculated as $\dot{p}_t = J(q_t)\dot{q}_t$.

\textbf{Collision avoidance constraint}: As mentioned in Section \ref{section:collision_detection}, we can get the $SDF$ for environmental collisions $e_t$ in real-time. The minimum distance vectors between robot and collisions $d_{static}$ and $d_{dynamic}$ are obtained given $q_t$. Thus, the env-collision constraint $C_{env\_col}$ is formulated as:
\vspace{-0.5em} 
\begin{equation}\label{eq:env_col_const}
C_{env\_col}(q_t,e_t,\epsilon_{env}) = \phi_1d_{static} + \phi_2d_{dynamic}- \epsilon_{env}
\vspace{-0.25em}
\end{equation}
Where the $\epsilon_{env}$ is the env-collision distance threshold, the $\phi_1$ and $\phi_2$ are weights. The self-collision is also considered by training a self-collision distance checking network as in \cite{bhardwaj2022storm}. The trained network gets input as $q_t$ and predicts the minimal distance between robot links, denoted as $d_{links}$. The self-collision constraint $C_{self\_col}$ is derived as:
\vspace{-0.5em} 
\begin{equation}\label{eq:self_col_const}
C_{self\_col}(q_t,\epsilon_{self}) = d_{links}-\epsilon_{self}
\vspace{-0.25em}
\end{equation}
Where the $\epsilon_{self}$ is the self-collision distance threshold. 
  
We solve the above optimization formulations by Model Predictive Path Integral (MPPI) Control refer to \cite{williams2017model}, and take the first point $q_1$ of the solved optimal trajectory $\zeta_{opt}$ for execution. The online time-optimal trajectory interpolation is employed inspired by~\cite{berscheid2021jerk} to further interpolate position commands for real-time position control of the robot.
\vspace{-0.25em}
\subsection{Task-level Planning Model}\label{sec:task}
\begin{figure}[t]
  \vspace{0.5em}
  \centering
  \includegraphics[width=0.40\textwidth]{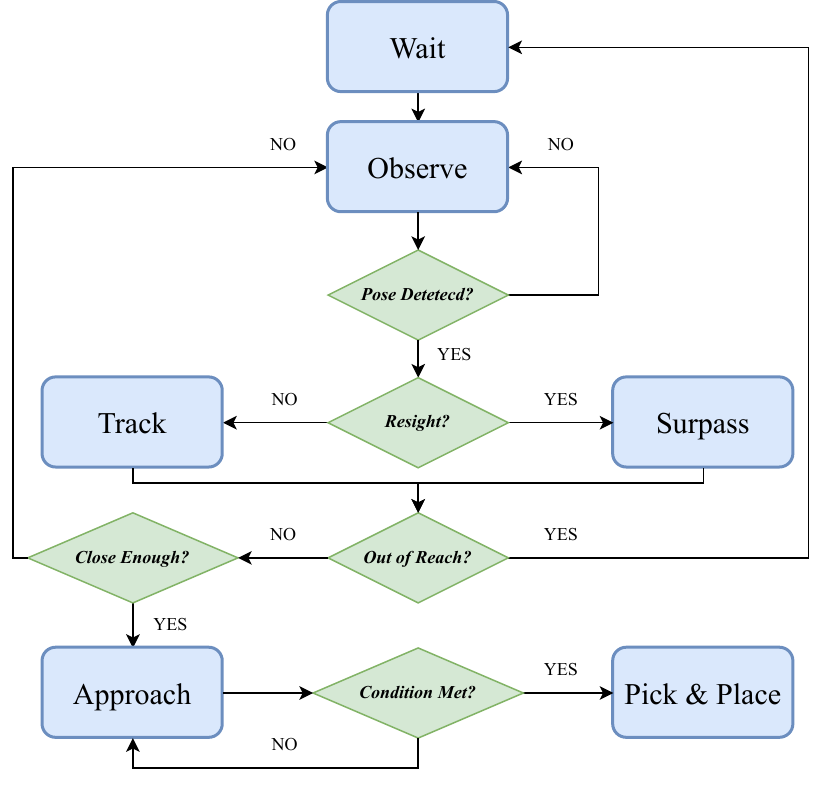}
  \caption{Task-level planning model consists of six actions for dynamic bin-picking. The actions are the Wait action in a standby state, the Observe action to perceive the suction pose and bin state, the Track action to follow the target pose, the Surpass action for rushing forward and regaining a decent view for observation, the Approach action to travel the final distance to contact with object surface, and the Pick \& Place action to attach the object from the bin and lift, and drop at placing location.}
  \label{fig:task_model}
  \vspace{-1.5em}
\end{figure}
The task-level planning model is shown in Fig. \ref{fig:task_model}. During the picking process, the robot initially waits for the bin to enter the perception region before determining the suction pose and bin state. However, challenges arise from ``Poor Observations" when the in-hand camera gets close to objects, including (a) occlusion causing insufficient point acquisition, (b) limited Field of View (FOV) missing points on the tracked object, and (c) perspective uncertainty from oblique views perceiving near-horizontal poses that are hard to execute. Thus, we developed a Resight policy (Alg.\ref{alg:resight}) that triggers a Surpass action to exceed the target and regain a better observation perspective. When the end-effector is within a distance $d_{thr}$ of the target pose, it starts approaching the object surface, following the velocity-matching objective. Upon detecting apparent contact force on the z-axis of the end-effector via the robot's built-in torque sensor, the vacuum generation is triggered. The contacted object is then picked, lifted, and dropped at a designated location.
\vspace{-0.75em}
\renewcommand{\thealgorithm}{1} 
    \begin{algorithm}
        \caption{Resight Policy} 
        \begin{algorithmic}[1]
            \Statex \textbf{Input:} Perception results from Observe action  
            \Statex \textbf{Output:} Track or Surpass action to be executed
            \State PCL Points threshold $N_{min}=1024$, Shift threshold $\epsilon_{shift}=0.05$, TAM threshold $M_{min}=0.3$, Scale-up factor $\sigma_{scale}=1.3$
            \State $ M \gets w_{1}T(p_t,\frac{r(s_{opt})-r(p_t)}{|r(s_{opt})-r(p_t)|}+w_{2}T(s_{opt},\frac{v_{ref}(s_{opt})}{|v_{ref}(s_{opt})|})$
            \If {$N_{pcl} > N_{min}$} {\color{blue}\Comment{No Occlusion}}
               \If {$\Vert r(s_{opt})-r(s_{t-1}) \Vert_2 < \vec{r} + \epsilon_{shift}$}
               \Statex{\color{blue}\Comment{No FOV Limitation}}
                  \If {$M> M_{min}$}{\color{blue}\Comment{No Oblique View}}\\
               \hspace{4em} $s_{tar} \gets s_{opt}+\vec{r}$
              \State \Return $q_{track} \gets \text{Optimization($q_t,s_{tar}$)} $
            \EndIf
            \EndIf
            \EndIf\\
             $s_{tar} \gets (s_{t-1}+ \sigma_{scale} \vec{r})+\vec{r} $
            \State \Return $q_{surpass} \gets \text{Optimization($q_t,s_{tar}$)} $
        \end{algorithmic}
        \label{alg:resight}
    \end{algorithm}
    \vspace{-1.25em}

\section{EXPERIMENTAL VALIDATIONS}
This section quantitatively evaluates the proposed DBPF for dynamic bin-picking tasks through several real-world experiments. First, we introduce the experiment setup and evaluation metric. Then, the ablation study is conducted. Followed by the comparisons of the proposed method with baseline methods are analyzed. Finally, The performance of our methods with varying characteristic variables of the fully dynamic scene is explored. 
\vspace{-0.25em}
\subsection{Experiments Setup and Evaluation Metrics}
\begin{figure*}[t]
    \vspace{0.5em}
	\centering
	\includegraphics[width=0.85\linewidth]{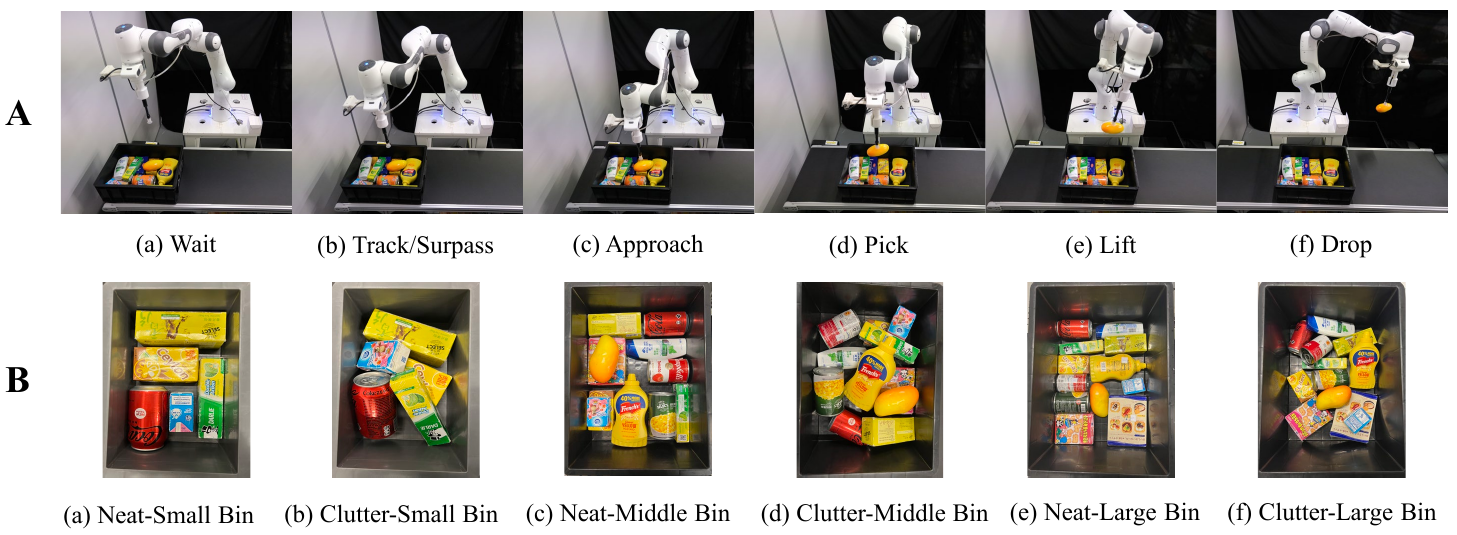}
    \vspace{-0.75em}
	\caption{Snapshots of dynamic bin-picking. A: (a)-(f) demonstrate the process of picking an object from a clutter of objects from a moving bin on a belt conveyor. B: (a)-(f) depict Neat and Clutter arrangements of objects in varying bins ranging from Small to Large in size. }
\label{fig:snap}
\end{figure*}

\begin{table}[h]
\vspace{-0.5em}
\caption{Ablation Experiments of The Proposed Framework On Dynamic Bin-picking}
\vspace{-1em}
\label{tab:ablation}
\begin{center}
\renewcommand\arraystretch{0.9}
    \begin{tabular}{cccc}
        \toprule
        \multirow{2}{*}{Methods} & \multicolumn{3}{c}{11 objects} \\
        \cmidrule(lr){2-4} 
        & SR  & TT (s) & CR  \\
        \cmidrule(lr){1-4}
        Ours-no-VM & 34\% & 4.7$\pm$1.0 & 0.0  \\
        Ours-no-RS & 67\% & 6.3$\pm$3.4 & 0.0 \\
        Ours-no-DO & 79\% & \textbf{3.9$\pm$1.3} & 19\%  \\
        Ours-no-TAM & 61\% & 4.7$\pm$1.0 & 0.0   \\
        Ours-Full & \textbf{87\%} & 6.6$\pm$2.1 & 0.0  \\
        \bottomrule
    \end{tabular}
\end{center}    
\vspace{-1.5em}
\end{table}

\textbf{Experiments Setup.} A standard logistic container bin containing several arbitrary objects is moved on a belt conveyor or disturbed by human hands. A 7-DOF Franka Panda manipulator equipped with a suction gripper system is employed to generate vacuum force on objects' surfaces for picking. A Realsense D435 camera is mounted on the Franka wrist for object perception. A third-view fixed camera Azure Kinect is placed externally, observing the environment's dynamic collision, i.e., the moving bin. 
For the sake of precise and timely bin collision perception, we use ArUco fiducial markers~\cite{garrido2014automatic} placed on one wall of the bin and a location with known transformation to the robot base. ROS is employed to enable seamless communication between all the modules.

\textbf{Evaluation Metrics.} We take the \textit{Success Rate} (SR) - The percentage of successful trials concerning total trials, in which the success trial is defined as successfully picking the object from the moving bin; \textit{Total Time} (TT) - Time in seconds elapsed to finish picking motion; \textit{Collision Ratio} (CR) - The ratio of collisions over the total trials during the whole picking procedure. We
implement 100 trials on each scenario setting for every method.
\vspace{-0.25em}
\subsection{Ablation Study}
We conducted ablation studies to evaluate the performance of novel designs in our framework. We implement the ablations: (1) \textit{Ours-no-VM}: the velocity-matching is excluded, replacing by open-loop approaching based on predicted pose; (2) \textit{Ours-no-RS}: the resight policy is discarded, turning to execute Track action always (3) \textit{Ours-no-DO}: the dynamic obstacle detection and avoidance are disabled; (4) \textit{Ours-no-TAM}: \textcolor{black}{the TAM score is not considered during pose selection;} (5) \textit{Ours-Full}: the complete version of our DBPF. 11 objects are placed in a moving bin with a speed of 8cm/s. 

The results are shown in Table \ref{tab:ablation}, \textit{Ours-no-VM} achieves the lowest success rate of 34\% compared to \textit{Ours-Full} of 87\%. The disparity reveals the significance of velocity matching. It is more effective to pick up the object by maintaining relative still, as this facilitates the suction cup seal formation on the object's surface. In addition, the velocity matching during the approach action is implemented in a velocity-feedback closed-loop fashion, which should be more robust than the open-loop approach that only relies on initial prediction.
\textit{Ours-no-DO} achieves the shortest total picking time but collides with the bin wall, as it excludes the computationally intensive collision checking.
\textit{Ours-no-RS} has almost a 67\% success rate showing a noticeable drop to \textit{Ours-Full}. 
The robot often struggles with ``Poor Observations" accompanied by the in-hand camera configuration during object tracking. The result indicates the effect of the resight policy, which helps the robot regain a better perspective to perceive more feasible picking poses. \textit{Ours-no-TAM} exhibits a considerable decrease in success rate, with a degradation of 29.5\% compared to \textit{Ours-Full}. The absence of manipulability consideration in \textit{Ours-no-TAM} leads to singularities in the robot's motion and are disadvantageous for continuous moving along with moving target, thereby causing picking failures.

\begin{table*}
\vspace{-0.75em}
\vspace{0.5ex}
\renewcommand\arraystretch{0.85}
    \centering
    \caption{Baseline Comparison Experiments with Proposed Framework in Dynamic Bin-picking Tasks}
    \vspace{-0.5em}
    \begin{tabular}{ccccccccccc}
        \toprule
        \multirow{2}{*}{Scene} & \multirow{2}{*}{Methods} & \multicolumn{3}{c}{1 object} & \multicolumn{3}{c}{5 objects} & \multicolumn{3}{c}{11 objects} \cr
        \cmidrule(lr){3-5} \cmidrule(lr){6-8} \cmidrule(lr){9-11}
        & & SR  & TT (s) & CR & SR  & TT (s) & CR & SR  & TT (s) & CR  \cr
        \cmidrule(lr){1-11}
        \multirow{1}{*}{Dynamic-to-static}
        & SPA & 96\% & 12.3$\pm$0.7 &0.0 & 95\% & 13.9$\pm$1.2 & 0.0 & 96\% & 13.0$\pm$0.8 & 0.0 \cr
        \midrule
        \multirow{4}{*}{Fully dynamic}
        & SPA & 5\% & 5.6$\pm$0.4 & 69\%  & 19\% & 6.0$\pm$0.5 & 71\% & 14\% & 5.5$\pm$0.4 & 68\%\cr
        & PBVS & 14\% & \textbf{4.7$\pm$1.3} & 33\% & 30\% & \textbf{5.5$\pm$0.9} & 44\% & 28\% & \textbf{5.1$\pm$1.0}  & 27\% \cr
        & LSPPA& 57\% & 7.2$\pm$3.9 & 21\% & 63\% & 5.8$\pm$0.6 & 14\% & 59\% & 5.8$\pm$1.6 & 17\% \cr
        & Ours &  \textbf{89}\% &  5.0$\pm$1.0 &  \textbf{0.0} & \textbf{81}\% & 5.5$\pm$2.3 & \textbf{0.0} & \textbf{80}\% & 5.2$\pm$1.3 & \textbf{0.0} \cr
        \midrule
        \multirow{4}{*}{Disturbed static}
        & SPA & 37\% & 7.4$\pm$0.7 & 39\% & 71\% & 8.1$\pm$0.7 & 43\% & 62\% & 8.7$\pm$1.7 &42\% \cr
        & PBVS & 36\% &  \textbf{5.4$\pm$0.4} & 30\% & 59\% &  \textbf{5.3$\pm$0.4} & 29\% & 64\% &  \textbf{5.2$\pm$0.3} & 29\% \cr
        & LSPPA & 75\% & 5.6$\pm$0.7 & 18\% & 72\% & 6.1$\pm$1.0 & 12\% & 64\% & 6.3$\pm$0.9 & 14\%  \cr
        & Ours & \textbf{83}\% & 7.5$\pm$3.0 & \textbf{0.0} & \textbf{85}\% & 8.0$\pm$3.7 & \textbf{0.0} & \textbf{87}\% & 6.9$\pm$3.1 & \textbf{0.0} \cr
        \bottomrule
    \end{tabular}\vspace{0cm}
    \label{tab:performance}
    \vspace{-2em}
\end{table*}
\vspace{-0.25em}
\subsection{Comparison to Baseline methods}
 We evaluate our framework against three potentially competitive baselines. For fairness, we slightly modify these baselines in object perception by using our suction pose generation module, and make sure the same maximum joint velocity of the robot during motion:

\noindent\textbf{Sense-Plan-Act~(SPA)}: The picking pose is generated once at the beginning, and offline motion planning is conducted based on the classic sampling planner RRT~\cite{kuffner2000rrt} considering pre-defined static collision, and then the robot is triggered to execute the planned trajectory towards the target object. This method is widely applied in existing static robotic bin-picking scenes. 

\noindent\textbf{Position-based Visual Servoing~(PBVS)}: The robot continuously tracks the perceived picking pose through position-based visual servoing~\cite{de2021dual}. When the tracking is close to the target pose, an open-loop approaching is conducted for the final distance.

\noindent\textbf{Loop-Sense-Predict-Plan-Act~(LSPPA)}: We adapted the method from Akinola et al.\cite{akinola2021dynamic} as a baseline, designed for grasping a single moving object on a belt conveyor. Given the newly generated picking pose, the motion trajectory is planned using the PRM\cite{kavraki1996probabilistic} planner with online trajectory seeding and static collision avoidance. The system implements trajectory blending for each new trajectory during execution and approaches a predicted pose in an open-loop manner when close enough. We adapted our TAMN to handle multiple objects, replacing their consideration of reachability.

In the following, we implement several experiments to evaluate the performance of our DBPF \textbf{(Ours)} against the above-mentioned baselines. The experiments are conducted in three scenes: (i) \textit{Dynamic-to-Static}: This represents the classic industrial scene, in which the bin is transported in front of the robot and stopped. The bin-picking starts in a static situation. (ii) \textit{Fully Dynamic}: The robot picks objects while the bin moves continuously. The bin moves linearly with a constant speed of 6cm/s in all settings. (iii) \textit{Disturbed Static}: The bin is stationary in the picking area of the robot. After the robot starts picking motion, the bin is disturbed manually by linear displacements in the \textit{xy} plane. In addition, we vary the number of arbitrary objects packed in the bin to 1, 5, and 11 objects, respectively. 

\begin{table*}
    \vspace{0.5em}
    \renewcommand\arraystretch{0.85}
    \centering
    \caption{Performance of Proposed Framework with Varying Fully Dynamic Characteristic Variables}
    \vspace{-0.5em}
    \begin{tabular}{ccccccccccc}
        \toprule
        \multirow{2}{*}{Arrangement} & \multirow{2}{*}{Conveyor speed} & \multicolumn{3}{c}{Small Bin } & \multicolumn{3}{c}{Middle Bin} & \multicolumn{3}{c}{Large Bin} \cr
        \cmidrule(lr){3-5} \cmidrule(lr){6-8} \cmidrule(lr){9-11}
        & & SR  & TT (s) & CR & SR  & TT (s) & CR & SR  & TT (s) & CR  \cr
        \cmidrule(lr){1-11}
        \multirow{4}{*}{Neat}
        & Slow & 87\% & 7.3$\pm$4.0 & 0.0 & 80\% & 5.6$\pm$4.1 & 0.0 & 78\% & 7.3$\pm$3.4 & 0.0 \cr
        & Middle & 77\% & 7.8$\pm$6.9 & 0.0 & 90\% & 7.4$\pm$1.0 &0.0 & 81\% & 6.7$\pm$1.9 & 0.0 \cr
        & Fast & 81\% & 7.7$\pm$3.7 & 0.0 & 74\% & 7.0$\pm$6.2 & 0.0 & 79\% & 7.0$\pm$3.2 & 0.0 \cr
        & Varing & 73\% & 7.6$\pm$4.6 & 0.0 & 72\% & 6.7$\pm$2.6 & 0.0 & 70\% & 7.3$\pm$4.4 & 0.0 \cr
        \midrule
        \multirow{4}{*}{Cluttered}
        & Slow & 74\% & 6.4$\pm$3.9 & 0.0 & 83\% & 5.3$\pm$1.7 & 0.0 & 73\% & 7.2$\pm$3.5 & 0.0 \cr
        & Middle & 72\% & 8.3$\pm$4.4 & 0.0 & 88\% & 6.6$\pm$2.1 & 0.0 & 73\% & 7.1$\pm$3.9 & 0.0 \cr
        & Fast & 80\% & 7.0$\pm$7.0 & 0.0 & 71\% & 6.9$\pm$1.7 & 0.0 & 81\% & 7.3$\pm$3.7 & 0.0 \cr
        & Varing & 81\% & 7.6$\pm$6.2 & 0.0 & 73\% & 7.6$\pm$8.4 & 0.0 & 74\% & 9.4$\pm$7.0 &0.0\cr
        \bottomrule
    \end{tabular}\vspace{0cm}
    \label{tab:scenevariable}
    \vspace{-1.75em}
    \end{table*}
    
Fig.\ref{fig:snap}A shows experiment snapshots of picking an object from the moving bin using our DBPF. The obtained results are summarised in Table \ref{tab:performance}. 
Traditional SPA takes nearly 13 seconds to pick up an object from the bin in the \textit{Dynamic-to-Static} scene, counting the time spent before the bin is stopped. In comparison, our DBPF significantly reduces the picking time to approximately 6 seconds while maintaining a slightly lower success rate, showcasing the significant improvement in picking efficiency enabled by our approach. 

In the \textit{Fully Dynamic} scene, our method displays a superior overall success rate of approximately 84\%, exceeding the highest performing baseline (LSPPA) which achieves a success rate of 60\%. The PBVS has the quickest picking time due to the approximate linear motion on the end-effector for the visual-servoing method, but it also has the second-lowest success rate. The SPA method exhibits the lowest success rate and highest collision ratio, attributable to its open-loop workflow. Remarkably, our method achieves zero collisions during picking, contrasting sharply with the SPA, which has the highest collision ratio at 70\%. This can be attributed to the following reasons: 1) Our pose selection with the proposed TAMN filters out hard-to-execute or inconsistent picking poses. Instead, PBVS and LSPPA often fail due to jittery motion and oscillation because of frequently changing picking poses; 2) Our use of velocity matching ensures relative stillness and closed-loop approaching, enhancing the success rate over open-loop methods used in the baselines. This aids in cup seal formation and adapts to dynamics once the approaching begins; 3) Our resight policy handles unexpected ``Poor Observations" situations frequently encountered in dynamic bin-picking, which causes the loss of tracking for baseline methods leading to picking failures; 4) The consideration of dynamic obstacles, ensures the safety and reliable operation throughout the process, which is proven to be collision-avoidance-effective in experiments.

\begin{figure}[htbp]
  \vspace{0.75ex}
  \centering
  \includegraphics[width=0.98\linewidth]{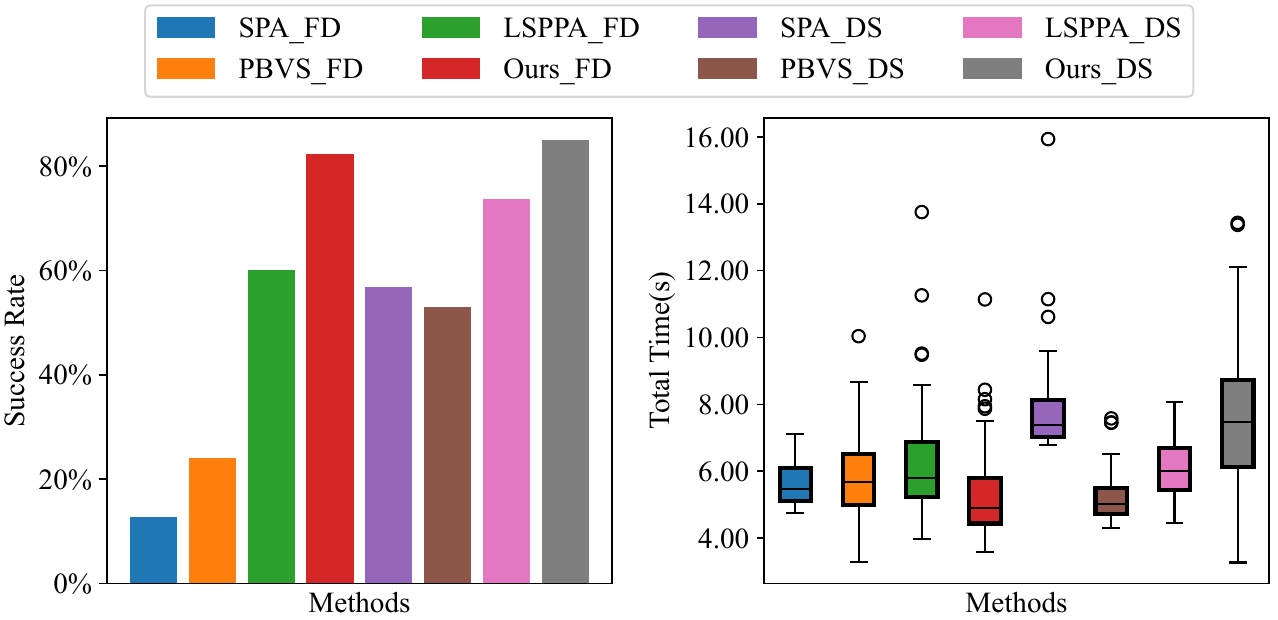}
  \vspace{-0.75em}
  \caption{Success rate (left) and total time (right) of dynamic bin-picking. *\_FD: Methods under the Fully Dynamic scene. *\_DS: Methods under the Disturbed Static scene. }
  \label{box_fig}
  \vspace{-2.0em}
\end{figure}

As for the \textit{Disturbed Static} scene, our method continues to show the highest success rate over the baselines when faced with external disturbances, as illustrated in Fig. \ref{box_fig} (Left). This property is crucial since many current static bin-picking applications typically assume completely controlled environments and disregard potential disturbances during picking. However, our method does exhibit a relatively high time variance under \textit{Disturbed Static} scene compared to other baselines in Fig. \ref{box_fig} (Right). This is primarily due to the random disturbances that can easily lead to "Poor Observations." Our resight policy helps to trigger the Surpass action, adjusting the camera's perspective and converging the tracking to a more desirable state before approaching, albeit with a longer duration. Besides, changing collisions can disrupt the original tracking motion, requiring more time to avoid collisions. Despite these challenges, our experimental results demonstrate that our framework surpasses baseline methods in dynamic bin-picking scenes, providing expected efficiency and reliability.
\vspace{-0.5em}
\subsection{Evaluation on varying Fully Dynamic scenarios}
This section explores our framework's potential for the \textit{Fully Dynamic} scenes with varying characteristic variables. We define two object arrangements: \textit{Neat} and \textit{Clutter}. In the \textit{Neat} arrangement, arbitrary objects in the bin are organized with minimal rotations and negligible overlap between objects. Conversely, the \textit{Clutter} arrangement sees objects randomly placed in the bin, in contrast to the \textit{Neat} arrangement. For each arrangement (Fig. \ref{fig:snap}B), we use bins with different sizes, i.e., \textit{Small} of $30\times20\times12$ (cm), \textit{Middle} of $41\times30\times15$ (cm), and \textit{Large} of $48\times35\times22$ (cm). Additionally, the bin moves at four different conveyor speeds: \textit{Slow} at 6cm/s, \textit{Middle} at 8cm/s, \textit{Fast} at 10cm/s, and \textit{Varying}, where the speed changes between 6cm/s to 10cm/s. 

Referring to the experimental results (Table \ref{tab:scenevariable}), our framework displayed comparable success rates in both \textit{Neat} and \textit{Clutter} arrangements, emphasizing DBPF's robustness across different levels of object arrangement complexity. For varying bin sizes, the success rate was found to be lower in the \textit{Large} bin compared to the \textit{Small} or \textit{Middle} bins. This is primarily due to the increased impact of the larger bin walls on the robot's motion, as collisions can disrupt normal picking actions. As for conveyor speed variation, it appears to have a minimal impact on our method. However, we noted a slight decrease in success rates under \textit{Varying} speed conditions compared to constant speed conditions. The primary reason for this is that changing speeds increase the dynamism of the environment, consequently escalating perception demands and necessitating a higher refresh rate. This could lead to delays in pose determination and bin state prediction, yielding a higher rate of picking failures. Lastly, the robot exhibits a shorter picking time when the bin moves at a \textit{Slow} speed. This is expected because minor environmental changes make tracking actions easier to converge, thereby facilitating a quicker approach to the object.
\vspace{-0.5em}
\textcolor{black}{\subsection{Failure case and Limitations}
\vspace{-0.25em}
Our system has certain limitations. Deformable objects hinder suction grasping due to poor sealing. Exclusively considering the linear motion, pick failures occur when bins or objects get rotated. Requiring prior knowledge of the environment, like bin model and fiducial markers, reduces system flexibility. Reflective and transparent objects cause low-quality point cloud, adversely affecting grasp pose decisions. With the need for obstacle avoidance, picking near bin walls encumbers the swift picking motion. The unknown relative pose between the gripper and picked object impedes further manipulation, due to scene-level pose generation.}

\textcolor{black}{\section{CONCLUSIONS AND FUTURE WORK}
We introduced a novel and efficient framework for dynamic bin-picking, efficiently identifying optimal suction poses for moving objects and leveraging horizon-based discrete trajectory optimization for reactive motion control and dynamic obstacle avoidance. 
Future work includes replacing the suction gripper by a two-fingered gripper with collision modeling and grasp pose generation network. Also, we will address current limitations such as considering the rotational motion of bins and objects, directly handling unknown environments via learning-based methods, and improving the post-picking process by adding an in-hand pose estimation module for subsequent precise placement or assembly tasks.} 
\vspace{-1.0em}

\bibliographystyle{IEEEtran}
\bibliography{ref}

\end{document}